\documentclass[a4paper, 10pt, conference]{ieeeconf}

\IEEEoverridecommandlockouts

\usepackage{graphics}
\usepackage{epsfig}

\usepackage{amsmath}
\usepackage{amssymb}
\usepackage{float}
\usepackage{physics}
\usepackage{siunitx}

\usepackage{blindtext}
\usepackage{hyperref}
\hypersetup{colorlinks, 
            linkcolor=blue,
            citecolor=blue,
            urlcolor=red}  

\usepackage{lipsum}

\usepackage{xcolor}
\definecolor{BrickRed}{HTML}{B6321C}

\newcommand{\nuc}{\newcommand}
\nuc{\clr}{\color{BrickRed}}

\usepackage{array}
\newcolumntype{P}[1]{>{\centering\arraybackslash}p{#1}}
\newcolumntype{L}[1]{>{\arraybackslash}p{#1}}
\usepackage{multirow}
\usepackage{subcaption}

\usepackage{dirtree}

\usepackage{balance}

\usepackage{csquotes}
\usepackage{acronym}

\usepackage[url=false, doi=false, isbn=false, sorting=none]{biblatex}
\title{\LARGE \bf

Feeling the Force: A Nuanced Physics-based Traversability Sensor for Navigation in Unstructured Vegetation
}

\author{Zaar Khizar$^{1,3}$, Johann Laconte$^{2}$, Roland Lenain$^{2}$ and Romuald Aufrère$^{3}$
\thanks{$^{1}$ LIMOS, Université Clermont Auvergne,  Clermont Auvergne INP, CNRS,  F-63000 Clermont-Ferrand, France}%
\thanks{$^{2}$ Université Clermont Auvergne, INRAE, UR TSCF, 63000, Clermont-Ferrand, France }%
\thanks{$^{3}$ Institut Pascal, Université Clermont Auvergne, Clermont Auvergne INP, CNRS,  F-63000 Clermont-Ferrand, France}%
}

\begin{document}
\hyphenpenalty=10000
\exhyphenpenalty=10000

\maketitle
\thispagestyle{empty}
\pagestyle{empty}

\begin{abstract} \label{sec:abstract}
In many applications, robots are increasingly deployed in unstructured and natural environments where they encounter various types of vegetation. 
Vegetation presents unique challenges as a traversable obstacle, where the mechanical properties of the plants can influence whether a robot can safely collide with and overcome the obstacle. 
A more nuanced approach is required to assess the safety and traversability of these obstacles, as collisions can sometimes be safe and necessary for navigating through dense or unavoidable vegetation. 
This paper introduces a novel sensor designed to directly measure the applied forces exerted by vegetation on a robot: 
by directly capturing the push-back forces, our sensor provides a detailed understanding of the interactions between the robot and its surroundings. 
We demonstrate the sensor's effectiveness through experimental validations, showcasing its ability to measure subtle force variations.
This force-based approach provides a quantifiable metric that can inform navigation decisions and serve as a foundation for developing future learning algorithms.
\end{abstract}

\section{INTRODUCTION}\label{sec:introdunction}

\begin{figure}[!t]
        \centering
        \includegraphics[width=\linewidth]{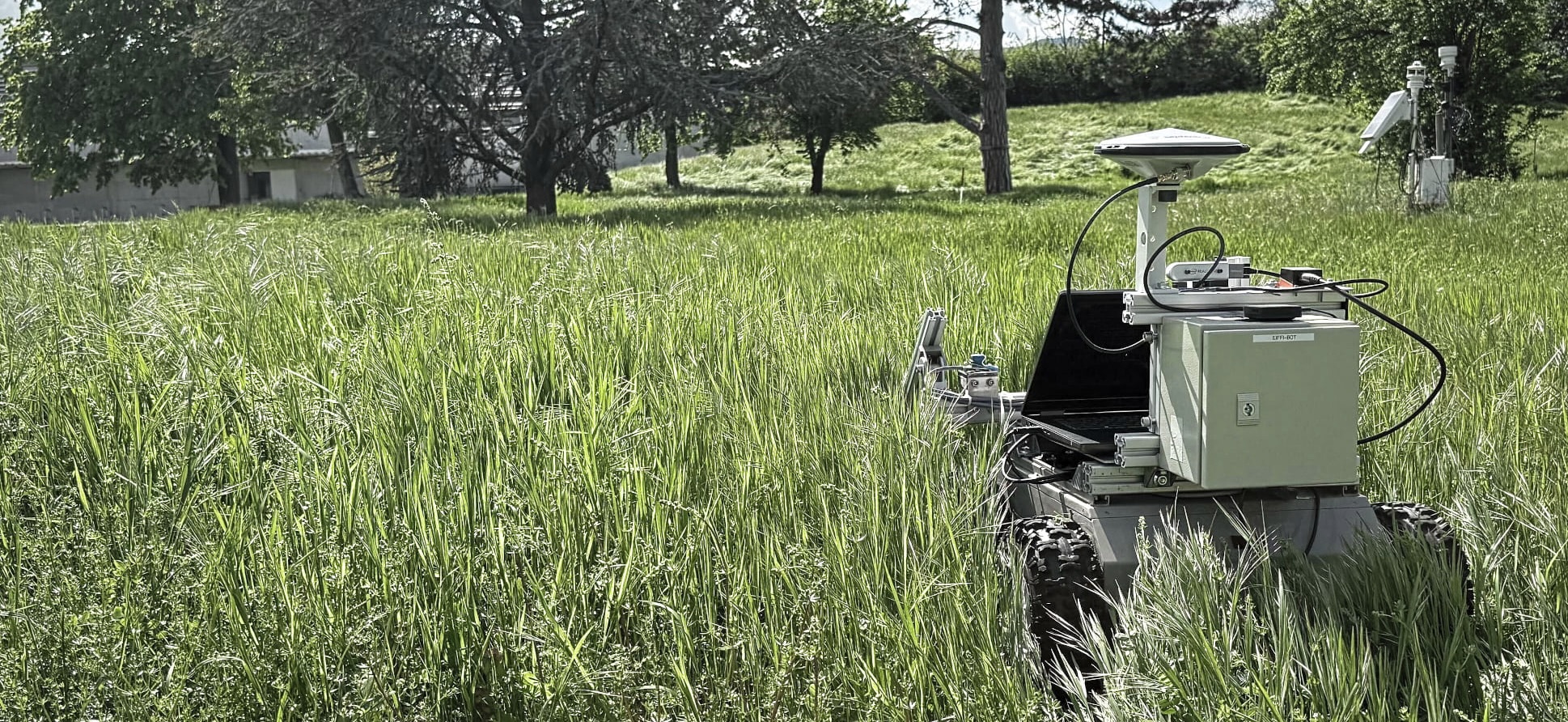}
        \caption{Robot navigating through unstructured environments. In such settings, collisions with vegetation are unavoidable. A nuanced approach to collision management is essential for effective navigation in these challenging environments.}
        \vspace{-0.75em}
        \label{fig:cover_photo}
    \end{figure}

    In unstructured natural environments, mobile robots frequently encounter dense vegetation, making contact with elements such as tall grass or low branches inevitable, as illustrated in \autoref{fig:cover_photo}. The ability to discern between obstacles that must be avoided and those that can be safely traversed is crucial in such settings. This distinction is particularly important in applications such as agriculture, forestry, or search and rescue, where robots operate in unstructured, natural terrain. While rigid obstacles generally pose a risk and are avoided, vegetation often allows safe physical interaction. However, conventional navigation systems tend to treat all obstacles conservatively, limiting the robot’s capacity to exploit the compliance of natural materials. This challenge highlights the non-trivial problem of managing collisions in such environments.

    Machine learning methods usually classify terrain traversability using visual and geometric cues. While effective in many settings, these methods typically lack an underlying model of physical interaction~\cite{frey2023fast}. This absence of a physical model makes it difficult to estimate the forces a robot might encounter in deformable terrain, as the methods rely solely on empirical data without considering the fundamental physics of the interaction. Consequently, their predictions may not accurately reflect the real-world dynamics or generalize well to new conditions and different robots~\cite{bradley2004vegetation}.
    
    This paper introduces a novel sensor and method for traversability analysis that directly measures robot-environment interaction forces. Unlike detailed deformation modeling in controlled environments \cite{frank2014learning}, our approach measures the force necessary for an obstacle to yield and allow traversal. A custom deformability sensor integrated into the robot captures impact forces, providing a physics-based metric for obstacle compliance that scales with robot and dynamics parameters. This enables informed decisions on traversing or avoiding regions.

    The main contributions of this paper are
    \begin{itemize}
        \item A novel sensor based framework for traversability analysis in deformable outdoor environments; and
        \item A real-time measurement system for capturing interaction forces during robot environment contact.
    \end{itemize} 

\section{Related Work}\label{sec:lit_review}

    Binary maps that classify space as either free or occupied have been foundational in structured environments \cite{sanchez2021path}. However, they fall short in natural settings where obstacles like vegetation exhibit varying deformability. Such simplifications lead to conservative behavior and reduced adaptability. Navigating these environments requires models that capture the dynamic, continuous nature of terrain, including physical interactions during contact.

\subsection{Modeling Traversability}
    Traversability modeling often relies on geometric, semantic, or learned representations, but rarely incorporates direct physical interaction, limiting fidelity in capturing terrain compliance.

    Laconte et al.~\cite{laconte2021novel} proposed the Lambda Field, a continuous occupancy map modeling collision intensity, supporting risk metrics for deformable elements like grass, though based on lidar inference, not measurement. Fan et al.~\cite{fan2021step} introduced STEP, a planner for extreme terrains, such as caves, using geometric and semantic cues with \ac{CVaR}, but lacking physical interaction modeling. Polevoy et al.~\cite{polevoy2022complex} predicted terrain disturbance from visual inputs by learning model error during execution, capturing uncertainty but not mechanical interaction. Cai et al.~\cite{cai2022risk} used deep learning to assess terrain traction, focusing on slip risk rather than vegetation compliance. Sathyamoorthy et al.~\cite{sathyamoorthy2023using} improved navigation in dense vegetation by classifying obstacle flexibility via lidar intensity, but relied on indirect visual cues.

    These approaches use exteroceptive sensing and diverse strategies, but none provide direct measurements of interaction forces.

\subsection{Deep Learning Based Approaches}
    Recent research in traversability analysis has increasingly adopted data driven methods using deep learning to handle the complexities of unstructured and deformable environments. These frameworks enable robots to learn terrain adaptability from multimodal sensor data, although modeling physical interactions remains a challenge.

    Gasparino et al. introduced WayFAST \cite{gasparino2022wayfast} and its successor WayFASTER \cite{gasparino2024wayfaster}, that are self-supervised frameworks fusing RGB-D data with traction estimates for traversability prediction, with WayFASTER improving robustness by generating temporally aware bird’s-eye-view maps to overcome the spatial limitations of image-based prediction.
    Frey et al.~\cite{frey2023fast} utilized self-supervised transformers for real-time forest navigation, though single camera constraints led to segmentation artifacts. Elnoor et al.~\cite{elnoor2024amcoadaptivemultimodalcoupling} fused vision-proprioception, but camera reliability suffered in low light conditions. Sathyamoorthy et al.~\cite{sathyamoorthy2023vernvegetationawarerobotnavigation} combined few shot learning with lidar, yet training bias affected generalization to novel plants. Sathyamoorthy et al.~\cite{sathyamoorthy2022terrapn} introduced TerraPN, leveraging self-supervised odometry signals for adaptive terrain costs, but odometry drift and computational demands persisted.
    
    While learning based methods improve terrain adaptability, they often rely on indirect signals, inferring compliance through visual or lidar data and estimating traversability via aggregated proprioceptive feedback. These methods face generalization, sensitivity, and computational challenges. In contrast, we propose a physically grounded deformability sensor that directly measures interaction forces with vegetation, offering a reliable complement to learning-based approaches.

\subsection{Modeling Deformability}
    Although deep learning effectively captures terrain patterns, its indirect sensing limits physical modeling essential for navigating deformable vegetation. Some studies address this using physics-aware models.
    
    Ahtiainen et al.~\cite{ahtiainen2015learned} fused radar and lidar to detect rigid objects in foliage, improving mapping but lacking adaptability and real-time performance due to offline training and computational cost. Frank et al.~\cite{frank2014learning} applied FEM with force-torque sensors and RGB-D for material learning, but were limited to predefined indoor objects and were computationally intensive. Haddeler et al.~\cite{haddeler2022traversability} used active terrain probing on legged robots, classifying surfaces by collapsibility thresholds. However, intermittent probing caused delays, and fixed thresholds struggled with graded vegetation deformability. Ordonez et al.~\cite{ordonez2018modeling,ordonez2020characterization} modeled vegetation as linearly deformable stems to estimate drag and energy cost, though generalization and real-time deployment remained limited.
    
    Al Bassit et al.~\cite{al2023sensitive} proposed the Sensitive Bumper Probing System, combining a load-cell bumper and lidar, which slows down to probe obstacles after detection. Goodin et al.~\cite{goodin2024measurement} embedded a force-sensitive pushbar in a ground vehicle, training a predictive model with synchronized force, lidar, and camera data. While these systems capture interaction forces, they are not sensitive enough for light vegetation such as grass or saplings. This is primarily because bumper based designs experience significant internal friction and structural damping, which can obscure the low magnitude forces exerted by flexible obstacles. In contrast, our custom wire-based sensor allows direct measurement of small interaction forces, making it more suitable for detecting and characterizing soft, deformable elements in vegetation rich environments.

\section{Methodology}\label{sec:Methodology}

    Accurate measurement of light deformable obstacles, such as grass, shrubs, and small branches, requires precise estimation of displacement and applied forces. To achieve this,  we employ a ratiometric potentiometer sensor, which measures the linear displacement of its wire. The wire is attached to an internal constant tension spring made of coiled metal strip, which provides a constant pullback force.
    
    \begin{figure}[H]
        \centering
        \includegraphics[width=0.9\linewidth]{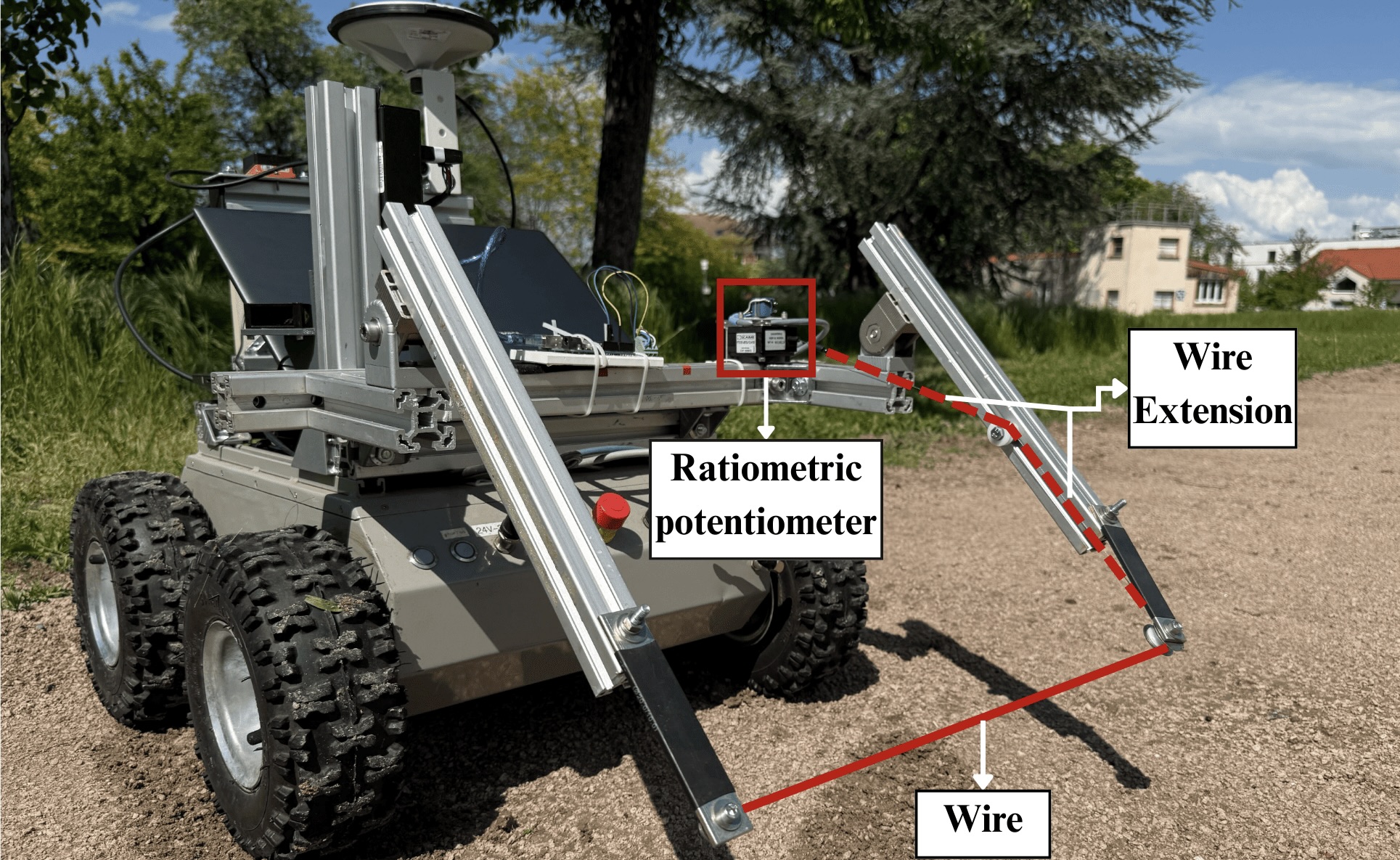}
        \caption{Our custom force sensor mounted on the robot, with the ratiometric potentiometer highlighted within a red box. The dotted red line indicates the part of the wire which extends from the sensor and the solid red wire indicates the part of the extension wire which comes in contact with the vegetation.}
        \vspace{-0.35em}
        \label{fig:sensor}
    \end{figure}
    
    The sensor is mounted on the robot, as illustrated in \autoref{fig:sensor}. A linear extension wire, mechanically linked to the sensor’s cable, is fixed to the robot and held under constant tension by the internal spring mechanism of the sensor. When the robot encounters an obstacle, the resulting force pushes the extension wire back toward the robot, causing additional wire to be pulled from the sensor's spool. Since the internal spring's constant force is known, the amount of wire pulled out can be directly related to the force applied, allowing us to estimate the vegetation resistive force.

\begin{figure}[!t]
        \centering
        \includegraphics[width=\linewidth]{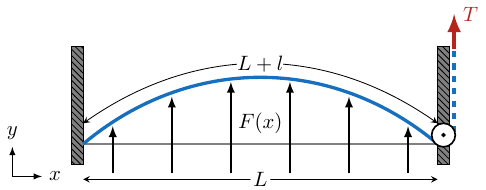}
        \caption{Schematic of the sensor model. The diagram illustrates a wire of rest length \(L\) that is extended by an additional length \(l\) under constant tension \(T\). The force distribution along the wire coming from the vegetation is given by \(F(x)\), where \(x\) represents the longitudinal distance, and the corresponding horizontal displacement is \(y(x)\).}
        \label{fig:wire_diagram}
        \vspace{-0.25em}
    \end{figure}

    We begin by describing the static equilibrium of our sensor system under the influence of forces exerted by environmental obstacles. \autoref{fig:wire_diagram} illustrates the schematic used in this analysis and defines the variables: 
    
    \begin{itemize}
        \item \(L\) denotes the rest length of the wire (solid red line in \autoref{fig:sensor}).
        \item \(l\) is the additional length pulled due to the applied force.
        \item \(T\) is the known constant tension in the wire.
        \item \(F(x)\) represents the force distribution of the vegetation along the wire.
        \item The relative displacement at position \(x\) is given by \(y(x)\), and its derivative and double derivative with respect to $x$ by \(y'(x), y''(x)\) respectively.
    \end{itemize}
    
    The potential energy of the system is composed of two parts. First, the potential energy stored in the wire, modeled as a constant-force spring, is given by
    \begin{equation}
        U_s = (l+L) \cdot T.
    \end{equation}
    Second, the potential energy due to the external force field (i.e., the vegetation) is expressed as
    \begin{equation}
        U_v = -\int_0^L F(x)\, y(x) \, \dd x.
    \end{equation}
    Thus, the total potential energy of the system is
    \begin{equation}
    \begin{aligned}
        U &= U_s + U_v = (l+L) \cdot T - \int_0^L F(x)\, y(x) \, \dd x \\
          &= \int_0^L \left[T\sqrt{1 + y'(x)^2} - F(x) y(x)\right] \dd x,
    \end{aligned}
    \label{eq:pot_en}
    \end{equation}
    where the term \(T\sqrt{1 + y'(x)^2}\) is the length of the deformed wire. 
    The static equilibrium is achieved when \(U\) is minimized.
    
    Given that the only known information is the pulled wire length \(l\), the generic case does not yield a unique solution from \autoref{eq:pot_en}.
    Therefore, we focus on two specific cases that represent typical interactions robot will encounter in natural settings: 
    \begin{enumerate}
        \item \textbf{Unique Displacement:} A discrete force acts at a specific point \(x=x_0\) on the wire, representing an interaction with a single thin object, such as a sapling.
        \item \textbf{Homogeneous Vegetation:} A constant force \(F(x)=F_v\) is applied along the wire, modeling interactions with uniform, dense vegetation.
    \end{enumerate}
    
    \subsection{Unique Displacement}
        In the case of a single object pushing on the wire at \(x=x_0\), the force is modeled as
        \begin{equation}
            F(x) = \delta(x - x_0) \cdot F_s,
        \end{equation}
        where \(\delta(x - x_0)\) is the Dirac delta function centered at \(x_0\), representing a concentrated force \(F_s\) applied at the longitudinal position \(x_0\in[0,L]\).
        The corresponding potential energy is given by
        \begin{equation}
            U = T \int_0^L \sqrt{1 + y'(x)^2}\,\dd x - F_s\, y_0.
        \end{equation}
        For \(x \in [0, x_0)\) and \(x \in (x_0, L]\), the Euler-Lagrange equation leads to straight line segments, as no external force acts on these portions of the wire. Since the wire is continuous, these two linear segments meet at \(x = x_0\), forming a triangular configuration with a peak height \(y_0 = y(x_0)\).

        The total length of the deformed wire corresponds to the sum of the two straight segments extending from the endpoints \(x = 0\) to \(x = x_0\) and from \(x = x_0\) to \(x = L\), each forming the sides of a right triangle.

        Thus, the total potential energy becomes
        \begin{equation}
            U = T \left(\sqrt{x_0^2 + y_0^2} + \sqrt{(L - x_0)^2 + y_0^2}\right) - F_s\, y_0.
        \end{equation}
        In this formulation, \(y_0\) is the only unknown as the positions \(x_0\) and \(L\) are fixed by the problem setup. 
        The shape of the wire and thus its potential energy depends solely on the height \(y_0\) at the point of contact.
        
        By setting the derivative \(\dv{U}{y_0}=0\), we obtain the force exerted by the object as
        \begin{equation}
        \begin{aligned}
            F_s &= T \left(\frac{y_0}{\sqrt{x_0^2 + y_0^2}} + \frac{y_0}{\sqrt{(L - x_0)^2 + y_0^2}}\right),
            \label{eq:force_single}
        \end{aligned}
        \end{equation}
        where \(y_0\) is determined by the constraint that the total length of the wire remains \(L+l\):
        \begin{equation}
            y_0=\frac{\sqrt{l(l + 2x_0)(l + 2L)\left(l + 2L - 2x_0\right)}}
                        {2(l + L)}.
        \end{equation}

        Now, consider the limit as the elongation increases, i.e., when \(l \to +\infty\). Physically, this corresponds to pulling the wire such that the point where the force is applied stretches further away, effectively stretching the two segments of the wire into nearly vertical orientations. 

        Therefore,  \(y_0 \to +\infty\) and in the limit of large elongation becomes
        \begin{equation}
            \lim_{y_0 \to +\infty} F_s = 2T.
            \label{eq:limit_unique}
        \end{equation}
        
        Beyond this point, no stable minimum of potential energy exists, as the system would continue to elongate indefinitely without restoring balance. As such, the maximum force the sensor in able to measure in the case of a single object is twice the tension of the ratiometric sensor.

    \subsection{Homogeneous Vegetation}
        For a homogeneous force field, the Lagrangian can be written as
        \begin{equation}
            \mathcal{L}(y, y') = T\sqrt{1 + y'(x)^2} - F_v\, y(x),
        \end{equation}
        and the potential energy to be minimized is
        \begin{equation}
            U = \int_0^L \mathcal{L}(y, y')\, \dd x.
        \end{equation}
        Using the Euler-Lagrange equation, we obtain
        \begin{equation}
        \begin{aligned}
            &-F_v - \dv{}{x}\left(\frac{T\,y'(x)}{\sqrt{1+y'(x)^2}}\right) = 0 \\
            \Leftrightarrow \quad F_v &= T\cdot\frac{y''(x)}{(1+y'(x)^2)^{3/2}} \\
            \Leftrightarrow \quad F_v &= T \cdot \kappa,
            \label{eq:force_homo}
        \end{aligned}
        \end{equation}
        where \(\kappa\) is the curvature of $y(x)$, the deformed wire. Thus, under homogeneous loading, \autoref{eq:force_homo} indicates that the curvature is constant and thus the wire assumes the shape of a circular segment.
        
        For a circular segment with central angle \(\theta\) and curvature \(\kappa\), the total deformed length of the wire is
        \begin{equation}
            L+l = \frac{\theta}{\kappa},
        \end{equation}
        and, by geometry,
        \begin{equation}
            L = \frac{2}{\kappa}\sin\left(\frac{\theta}{2}\right).
        \end{equation}
        Combining these expressions yields an equation for the curvature as a function of the length at rest $L$ and after deformation $L+l$:
        \begin{equation}
            L = \frac{2}{\kappa}\sin\left(\kappa\,\frac{L+l}{2}\right),
            \label{eq:curvature}
        \end{equation}
        which can be solved numerically for $\kappa$. In summary, given the wire displacement \(l\), the curvature \(\kappa\) is determined from \autoref{eq:curvature} and the homogeneous force is computed using \autoref{eq:force_homo}. Note that for a physically meaningful solution, the radius of the circular segment must satisfy \(2/\kappa \geq L\), implying that the maximum theoretical force is
        \begin{equation}
            F_v \leq \frac{2T}{L}.
            \label{eq:limit_homo}
        \end{equation}
        Above this limit, the whole wire would be under a total force of $L\cdot\frac{2T}{L}=2T$ and extend indefinitely before reaching its mechanical limit.
        
        In summary, \autoref{eq:curvature} and \autoref{eq:force_homo} provide the framework for computing the homogeneous force \(F_v\) from the measured wire displacement \(l\), while \autoref{eq:force_single} characterizes the force exerted by a discrete object at \(x = x_0\). These equations form the basis for interpreting sensor data under different interaction scenarios. Since only a single nonlinear equation must be solved per reading, computation is lightweight and well suited for real-time use.

\section{Experiments and Results}

        In this section, we present an experimental evaluation of the proposed traversability analysis framework. 
        These experiments are designed to capture the nuances of force interactions in varying natural terrains, showing the sensor's performance and the potential for integration with autonomous navigation strategies. 
        Three distinct experiments were performed: one in which the robot traverses a single sapling, another where it navigates through a patch of dense grass, and another where it traverses a shrub.
        
        The experiments were conducted on a mobile robot equipped with a ratiometric sensor mounted, as seen in \autoref{fig:sensor}, with the wire being at a height of \SI{0.16}{m} from the ground and \SI{0.44}{m} wide in length. 
        The wire of the sensor is maintained under a constant tension $T=\SI{2.2}{\N}$, thus being able to measure an overall force up to $2T=\SI{4.4}{\N}$. The wire sensor operates at a default sampling rate of 10\,Hz but, as it reads an analog signal, it is not inherently limited to this rate and can be configured for higher frequency sampling as needed. In addition, the pose of the robot is given via a fusion of an \ac{RTK GNSS} sensor, an \ac{IMU} as well as motor encoders. Throughout these experiments, the robot was manually controlled and the speed of the robot was fixed at \SI{1}{\m\per\s}. 

   \subsection{Experiment 1: Traversing a Single Sapling}
    
\begin{figure}[H]
            \centering
            \includegraphics[width=\linewidth]{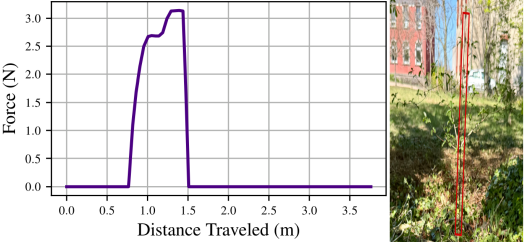}
            \caption{Force profiles and physical representation of a single sapling. \textit{Left:} Force profile as a function of the total distance travelled by the robot. \textit{Right:} Reference image of the actual sapling, highlighted in a red box for the ease of the reader.}
            \label{fig:sapling}
        \end{figure}

        This experiment investigates the robot’s interaction with a discrete deformable obstacle, a single sapling. The objective is to assess the sensor's ability to detect localized force events resulting from contact with a slender, vertical structure. In this setup, the robot was directed to traverse directly through the sapling, as seen in \autoref{fig:sapling}. The interaction was modeled as a single point force acting on the sensor’s wire at the midpoint \(x_0 = L/2\) for simplicity. While this assumption aligns with the rod-and-wire dynamics, which tend to converge the contact point toward the center under load, future work will explore using segmentation algorithms to detect the actual point of interaction along the wire and assess sensitivity to off-center impacts. The corresponding force was computed using the formulation presented in \autoref{eq:force_single}.
        
        The force–distance profile for the sapling interaction reveals two distinct phases of mechanical response. During the initial engagement at $[0.7,0.9]\,\SI{}{\m}$, the force increases nearly linearly from $[0.0,2.5]\,\SI{}{N}$, indicating elastic deformation of the sapling stem as it begins to bend. This is followed by a yield point phase, $[1.0,1.5]\,\SI{}{m}$, in which the force peaks at approximately $\SI{3}{N}$ around $\SI{1.2}{m}$, suggesting that the sapling has reached its elastic limit and begins to yield for the robot to traverse through it.

    \subsection{Experiment 2: Traversing patch of vegetation}
    
\begin{figure}[!ht]
            \centering
            \includegraphics[width=1\linewidth]{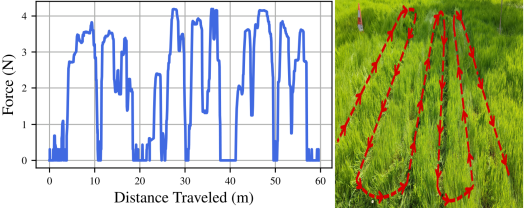}
            \caption{Force profile and physical representation of a dense grass patch. \textit{Left:} Force profile plotted against the cumulative distance travelled by the robot. \textit{Right:} Reference image showing the robot’s trajectory in red. }
            \label{fig:grass}
        \end{figure}

        In this experiment, the robot is deployed in the outdoor environment featuring a patch of dense grass, as seen in \autoref{fig:grass}, to assess the sensor's ability to detect homogeneous forces. As the robot traverses the patch, the force data is recorded using the \autoref{eq:force_homo}.
        
        The force profile reaches a maximum of $\SI{4.3}{N}$ and exhibits complex responses, reflecting spatial variations in grass density. Distinct peaks and troughs correspond to localized regions of denser vegetation. Force values in the range of  $[3.5,4.0]\,\SI{}{N}$ indicate areas of pronounced obstruction while values between $[0.0, 1.0]\,\SI{}{N}$ corresponds to the robot turning outside of the tall grass.
        
\begin{figure}[H]
          \centering
          \includegraphics[width=\linewidth]{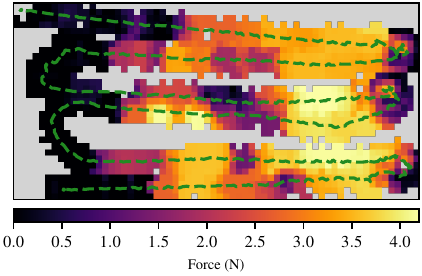}
          \caption{Force field map of the traversed grass patch. The environment is divided into $0.15\,\mathrm{m}$ cells, each colored by the mean force measured within that cell. Brighter colors indicate higher resistance, while darker colors indicate lower forces. The green dotted line shows the trajectory followed by the robot during traversal. Gray cells correspond to unexplored areas. }
          \label{fig:force-field-map}
        \end{figure}

        To complement these measurements with spatial context, we construct a force field map by discretizing the traversed area into a 2D grid with cell resolution of $0.15 \times 0.15\,\SI{}{\meter}$, as shown in~\autoref{fig:force-field-map}. At each time step, sampling at \SI{10}{Hz}, the sensor provides a single force reading representing the integrated interaction along the entire wire, which spans \SI{0.44}{\meter} laterally across the robot's front. This force is uniformly distributed along the length of the wire and projected into the grid by sampling intermediate points along the wire's arc, based on the robot's pose. For each sampled point, the corresponding grid cell is identified, and the measured force is accumulated into that cell. Over time, the average force per cell is computed by normalizing accumulated values by the number of contributions. The resulting map, overlaid on the robot’s trajectory, highlights local variations in vegetation density and terrain resistance, with brighter regions indicating higher forces and denser vegetation, and darker regions corresponding to lower resistance in sparser areas.

        By representing the environment as a continuous force field rather than a binary occupancy grid, the robot obtains a richer model of terrain compliance.         Consequently, path planning can distinguish between deformable obstacles that can be traversed and rigid obstacles that must be avoided, adapting routes according to both the robot’s force capabilities and the strength of environmental obstructions.

    \subsection{Experiment 3: Traversing a shrub}
    
\begin{figure}[H]
            \centering
            \includegraphics[width=\linewidth]{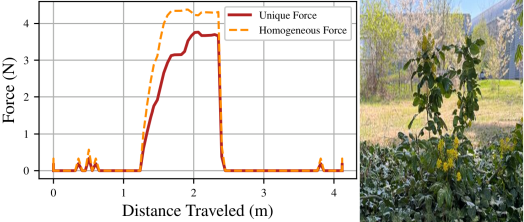}
            \caption{Comparative force profiles and physical representation of the shrub obstacle. \textit{Left:} Force–distance curves for the homogeneous model (\autoref{eq:force_homo}, orange) and the unique-displacement model (\autoref{eq:force_single}, red). \textit{Right:}  Reference image of the shrub structure.}
            \label{fig:shrub_comparison}
            \vspace{-0.5em}
        \end{figure}

        The current sensor implementation uses a fixed force model, limiting adaptability to obstacles with evolving contact characteristics. For example, when traversing a shrub, the force signal may be interpreted using either the homogeneous model~(\autoref{eq:force_homo}) for a smooth, averaged response or the unique‐displacement model~(\autoref{eq:force_single}) for discrete increments. Both yield similar peak forces of around $\SI{4}{N}$, as shown in~\autoref{fig:shrub_comparison}, though the system does not yet adaptively switch between models. Future work will incorporate exteroceptive sensors, such as RGB-D or lidar, to segment the wire and estimate contact locations in real time.  This will enable autonomous selection or switching between the two force models based on the observed interaction. These physical models are useful abstractions for most cases but remain idealizations. In more complex settings where vegetation is neither homogeneous nor reducible to a single stem, it may be necessary to approximate with one of the existing models or develop new formulations suited to the observed interaction profile.

        The proposed wire-based sensor reliably measures interaction forces up to approximately $4.4\,\mathrm{N}$, making it well suited for light vegetation such as dense grass, saplings, and small shrubs. Larger obstacles, such as mature shrubs or small trees, often exceed this threshold. To extend coverage, we plan to deploy an array of wire sensors with varying stiffness and integrate a bumper-based probing system similar to those in~\cite{al2023sensitive, goodin2024measurement}. While the wire may occasionally become entangled with vegetation, this is acceptable in data collection scenarios. However, such behavior would require mitigation if the sensor were to be deployed on an autonomous platform.

\section{Conclusion}

    This paper presented a sensor-based approach for characterizing deformable obstacles in unstructured natural terrains. By directly measuring interaction forces during traversal, the method complements vision-based perception by providing raw measurements of vegetation deformation forces. Field experiments demonstrated reliable force readings across different vegetation densities, supporting the sensor’s ability to capture terrain resistance. Sparse force field maps constructed from these readings offer a richer representation than binary occupancy grids, enabling planning that accounts for both deformable and rigid obstacles relative to the robot's force limits. Future work includes integrating a bumper system to extend sensing to larger obstacles, evaluating the effect of varying robot speeds, and systematically validating results against ground truth. We also plan to collect force data from diverse environments and explore learning based methods that fuse force, visual, and lidar inputs to predict dense, force-aware terrain maps.

\acrodef{CVaR}{Conditional Value at Risk}
\acrodef{RTK GNSS}{real-time Kinematic Global Navigation Satellite System}
\acrodef{IMU}{Inertial Measurement Unit}






\section*{ACKNOWLEDGMENT}
This work has been funded by the french National Research Agency (ANR) via the Junior Research Chair program.
This work was supported by the International Research Center “Innovation Transportation and Production Systems” of the I-SITE CAP 20-25.


\section*{REFERENCES}
\renewcommand*{\bibfont}{\footnotesize}
\printbibliography[heading=none]

\end{document}